\documentclass[10pt,twocolumn,letterpaper]{article}

\usepackage[pagenumbers]{cvpr} 
\usepackage[utf8]{inputenc}
\usepackage{listings}

\lstdefinelanguage{json}{
    basicstyle=\ttfamily\footnotesize,
    numbers=left,
    numberstyle=\tiny,
    stepnumber=1,
    numbersep=5pt,
    showstringspaces=false,
    breaklines=true,
    literate=
     *{0}{{{\color{black}0}}}{1}
      {1}{{{\color{black}1}}}{1}
      {2}{{{\color{black}2}}}{1}
      {3}{{{\color{black}3}}}{1}
      {4}{{{\color{black}4}}}{1}
      {5}{{{\color{black}5}}}{1}
      {6}{{{\color{black}6}}}{1}
      {7}{{{\color{black}7}}}{1}
      {8}{{{\color{black}8}}}{1}
      {9}{{{\color{black}9}}}{1}
      {:}{{{\color{black}{:}}}}{1}
      {,}{{{\color{black}{,}}}}{1}
      {"}{{{\color{red}{"}}}}{1},
}

\usepackage[T1]{fontenc}
\definecolor{cvprblue}{rgb}{0.21,0.49,0.74}
\usepackage[pagebackref,breaklinks,colorlinks,allcolors=cvprblue]{hyperref}
\usepackage{graphicx}
\usepackage{amsmath}
\usepackage{amssymb}
\usepackage{booktabs}
\usepackage{multirow}
\usepackage{pifont}

\usepackage[normalem]{ulem}
\useunder{\uline}{\ul}{}

\usepackage{colortbl} 
\usepackage{arydshln}
\usepackage[dvipsnames]{xcolor}
\usepackage{pifont} 
\usepackage{tikz} 
\definecolor{tabfirst}{rgb}{1, 0.75, 0.7}
\definecolor{tabsecond}{rgb}{1, 0.85, 0.65}
\definecolor{tabthird}{rgb}{1, 0.96, 0.7}

\def\paperID{2318} 
\def\confName{CVPR}
\def\confYear{2026}
\usepackage{colortbl}
\usepackage{arydshln}
\usepackage[dvipsnames]{xcolor}
\title{GaussianDWM: 3D Gaussian Driving World Model for Unified Scene Understanding and Multi-Modal Generation}

\author{ Tianchen Deng\textsuperscript{1}*, Xuefeng Chen\textsuperscript{2}*, Yi Chen\textsuperscript{1}*, Qu Chen\textsuperscript{\rm 3,4}, Yuyao Xu\textsuperscript{\rm 3,4}, Lijin Yang\textsuperscript{\rm 3,4}, \\ Le Xu\textsuperscript{\rm 3,4}, Yu Zhang\textsuperscript{\rm 3,4}, Bo Zhang\textsuperscript{\rm 3,4}, Wuxiong Huang\textsuperscript{\rm 3,4}, Hesheng Wang\textsuperscript{\rm 1}
 \\
{\textsuperscript{\rm 1} Shanghai Jiao Tong University}
{\textsuperscript{\rm 2} Tsinghua University}
{\textsuperscript{\rm 3} MEGVII Technology}
{\textsuperscript{\rm 4} Mach Drive}
}

\begin{document}

\twocolumn[{%
\renewcommand\twocolumn[1][]{#1}%
\maketitle

\begin{center}
  \centering
  \vspace{-18pt}
  \captionsetup{type=figure}
  \includegraphics[width=\linewidth]{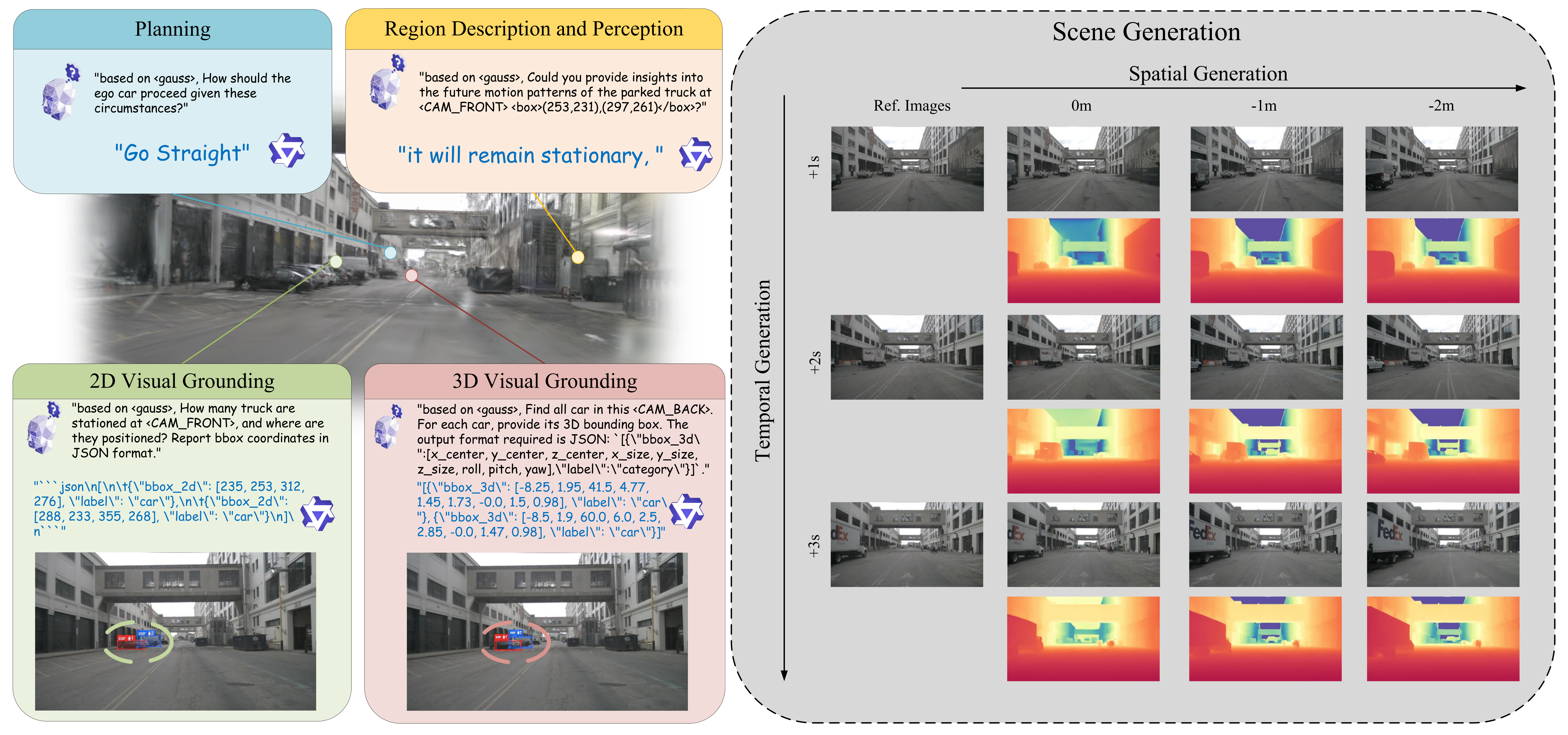}
  \setlength{\abovecaptionskip}{-2.5pt}
  \captionof{figure}{We propose the first unified 3D Gaussian-based world model framework 
that achieves comprehensive scene understanding and scene generation 
for driving scenarios. 
It efficiently encodes complex scenes, samples task-relevant information, 
and handles diverse question-answering tasks. 
Moreover, by leveraging the extracted world knowledge, 
our framework guides the generative model to perform accurate 
spatial and temporal scene generation.}
  \label{fig:teaser}
\end{center}%
}]

\renewcommand{\thefootnote}{} 
\footnotetext{ The first three authors contribute equally to this paper. Project leader: Qu Chen, Corresponding author: Hesheng Wang}

\begin{abstract}
Driving World Models (DWMs) have been developing rapidly with the advances of generative models. However, existing DWMs lack 3D scene understanding capabilities and can only generate content conditioned on input data, without the ability to interpret or reason about the driving environment. Moreover, current approaches represent 3D spatial information with point cloud or BEV features do not accurately align textual information with the underlying 3D scene. To address these limitations, we propose a novel unified DWM framework based on 3D Gaussian scene representation, which enables both 3D scene understanding and multi-modal scene generation, while also enabling contextual enrichment for understanding and generation tasks. Our approach directly aligns textual information with the 3D scene by embedding rich linguistic features into each Gaussian primitive, thereby achieving early modality alignment. In addition, we design a novel task-aware language-guided sampling strategy that removes redundant 3D Gaussians and injects accurate and compact 3D tokens into LLM.
Furthermore, we design a dual-condition multi-modal generation model, where the information captured by our vision-language model is leveraged as a high-level language condition in combination with a low-level image condition, jointly guiding the multi-modal generation process. We conduct comprehensive studies on the nuScenes, and NuInteract datasets to validate the effectiveness of our framework. Our method achieves state-of-the-art performance. We will release the code publicly on GitHub \href{https://github.com/dtc111111/GaussianDWM}{https://github.com/dtc111111/GaussianDWM} .
\end{abstract}

\section{Introduction}
Driving World Model (DWM)~\cite{vista,gaia,drivingwm,chen2026layer,xie2026raynova} have become essential for
autonomous driving for
their ability to predict future scene generation. These models predict environmental changes and synthesize simulation data for risk forecasting, route optimization, and corner case training. Most existing approaches achieve this by predicting modalities such as images~\cite{drivedreamer,TrajDiff} and point clouds~\cite{pointcloud,liang2025parameter,liang2024pointmamba}, which represent the visual and geometric properties of the environment. While these DWMs excel at forecasting how the environment may evolve, they are difficult to interpret, describe, or query, and cannot easily provide contextual information (e.g., visual question answering or scene description). With the rapid progress of Large-Language Models and Vision-Language Models (VLMs)~\cite{qwen3,wang2024eaco,wang2025care,zhu2025medeyes,wang2025selfdestructivelanguagemodel,wang2025reasoningretrievalstudyanswer}, remarkable advancements have been achieved in general vision tasks by leveraging world knowledge and causal reasoning. This highlights the potential of combining the scene understanding capability of VLMs with the generative power of DWMs as a promising future direction. Notably, pioneering efforts such as HERMES~\cite{hermes} and UniFuture~\cite{liang2025seeing} have first achieved the unification of scene generation and understanding for autonomous driving world models. They adopts BEV/Depth features to represent spatial information, aligns them with the text space, and incorporates them into the generative model. However, this BEV-based scene representation only achieves feature-level alignment between textual and spatial inputs, which is not sufficiently accurate. To overcome this limitation, we propose a novel 3DGS-based scene representation that unifies scene understanding and generation.

First, we directly embed linguistic features into the 3D Gaussians, thereby achieving explicit spatial alignment between text and the 3D scene. This improves the accuracy of cross-modal alignment. Second, given the redundancy of Gaussians (i.e., extremely dense representations with tens of thousands of tokens per scene), it is impractical for VLMs to process such a large number of tokens effectively. To address this, we introduce a novel task-aware language-guided sampling strategy. By calculating the similarity between the input text and the 3D Gaussians, our method selects  the most relevant Gaussians for the query and projects them into the context space through our projector, injecting accurate and compact 3D tokens into the textual understanding.
Finally, we design a dual-condition multi-modal generation model. In this framework, the understanding from the VLM provides a high-level language condition, while image features serve as a low-level image condition. Together, they guide the generation of multiple modalities, including RGB, depth, and language. Moreover, our framework supports both spatial and temporal generation.
Overall, our contributions are shown as follows:
\begin{itemize}
    \item We propose the first 3D Gaussian-based unified world model framework that supports both scene understanding and scene generation.
    \item We introduce a novel token extraction and projection module for 3D Gaussian scene representations. Due to the redundancy of 3D Gaussians, we further develop a task-aware language-guided sampling strategy that overcomes token length limitations while preserving essential spatial information. 
    \item We design a novel dual-condition multi-modal scene generation framework with high-level feature from world knowledge and low-level feature from images. 
    \item Extended experiments on the Nuscenes and NuInteract datasets demonstrate that our method effectively bridges the gap between understanding and generation, enabling both accurate scene comprehension and more coherent future scene prediction.
\end{itemize}
\section{Related Work}

\noindent\textbf{Novel View Synthesis for Urban Scene}
With the emergence of NeRF~\cite{nerf} and 3D Gaussian Splatting (3DGS)~\cite{3dgs}, many methods have adopted these representations across a wide range of tasks, including robotic mapping and localization~\cite{mneslam,mcnslam,plgslam,vpgsslam,deng2025omnimap,deng2024openobj}, VR~\cite{neslam,li2025human,gu2025mocount,zhu2025sni,gong2025dino,gong2025ov3r}, and autonomous driving~\cite{nsg,snlidar}. Early NeRF-based methods such as NSG~\cite{nsg}, SUDS~\cite{suds}, ProSGNeRF~\cite{deng2023prosgnerf}, EmerNeRF~\cite{emernerf}, and FreeDriveRF~\cite{freedriverf} achieved dynamic–static disentangled reconstruction through the use of scene graphs, optical flow, or other motion cues. More recently, several 3DGS-based approaches have been proposed to further improve rendering efficiency. PVG~\cite{pvg} introduces periodic vibration-based temporal dynamics to unify static and dynamic elements without manual annotations. Methods such as Street Gaussians~\cite{streetgaussian}, Driving Gaussian~\cite{drivinggaussian}, and DeSiRe-GS~\cite{DeSiRe-GS} also explicitly separate dynamic and static components for reconstruction. LESSON~\cite{lesson} proposes a teacher-guided diffusion strategy for generating 3D Gaussian splats using only 2D supervision. STORM~\cite{storm} proposes a feed-forward Transformer architecture to infer dynamic Gaussians and their velocities, enabling efficient large-scale outdoor scene reconstruction.  DrivingForward~\cite{drivingforward} achieves feed-forward reconstruction from sparse surround-view inputs using self-supervised pose and depth estimation. MUDG~\cite{mudg} and Dist-4D~\cite{dist4d} propose multi-modal novel view synthesis frameworks that generate both RGB and depth modalities.

\noindent\textbf{Driving World Model}
Driving world models~\cite{kong20253d,yan2025drivingsphere,tu2025role,linmore,fu2025orion} have attracted considerable attention in autonomous driving due to their ability to provide comprehensive environmental representations and predict future scenarios. Current methods primarily rely on 2D and 3D conditions for scene generation. GAIA-1~\cite{gaia1} introduces an autoregressive model for video generation in driving scenarios. DriveDreamer~\cite{drivedreamer} proposes a scene generation framework conditioned on 3D structure to provide geometric representations that benefit downstream autonomous driving tasks. MagicDrive~\cite{magicdrive} presents a street-view synthesis framework with precise 3D controls (e.g., camera poses, road maps) using cross-view attention, improving 3D object detection and BEV segmentation. DreamDrive~\cite{dreamdrive} further combines video diffusion with hybrid Gaussian scene representations to synthesize 4D scenes with 3D-consistent dynamic video rendering. Most recently, UniScene~\cite{uniscene} proposes an occupancy-centric approach that unifies semantic, visual, and LiDAR data generation. However, existing DWMs overlook the scene understanding ability of the driving environment.

\noindent\textbf{Large Language Models for Driving}
Large Language Models (LLMs) have demonstrated impressive generalization ability and extensive world knowledge across various tasks, including scene understanding~\cite{wang2025learning,3dllm,3dvla,zhang2025cross,zhu2025pathology}, visual question answering (VQA), and both 2D and 3D visual grounding. DriveGPT4~\cite{drivegpt4} processes front-view video inputs to predict vehicle actions while providing natural language justifications via an LLM. DriveLM~\cite{drivelm} leverages LLMs for graph-based VQA and end-to-end autonomous driving. 
NuInteract~\cite{drivemonkey} further integrates large vision-language models (LVLMs) with a spatial processor using a set of learnable queries, trained on the large-scale NuInteract dataset containing over 1.5M multi-view image–language pairs, covering dense scene captioning and diverse interactive tasks. GaussianVLM~\cite{gaussianvlm} introduces a 3D Gaussian-based 
visual question answering (VQA) framework, where a SceneSplat-style 
variational autoencoder (VAE) is employed to directly encode 3D Gaussian scenes. 
Hermes~\cite{hermes} proposes a BEV-based world model that integrates 
scene representation with a Vision-Language Model (VLM) 
for joint scene understanding and generation, 
which is the most closely related work to ours. 
However, existing approaches typically rely on image, point cloud, 
or BEV representations. 
In contrast, our method leverages a 3D Gaussian scene representation~\cite{deng2025best3dscenerepresentation} 
that achieves explicit spatial alignment between language features 
and 3D geometry, resulting in more accurate multimodal correspondence 
and improved representation of both texture and structural information 
in complex environments.

\begin{figure*}[h]
    \centering
    \includegraphics[width=\linewidth]{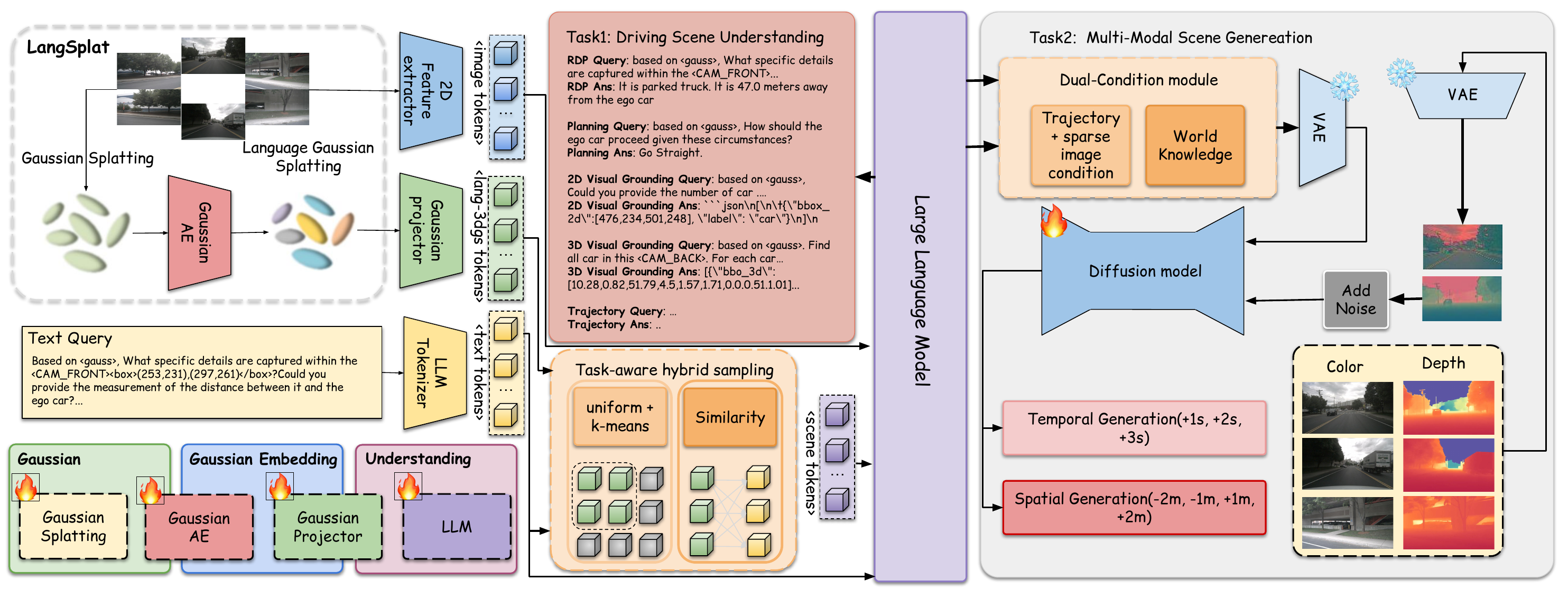}
    \vspace{-0.2cm}
    \caption{\textbf{System Overview.} We propose the first unified 3D Gaussian-based world model framework that simultaneously supports both scene understanding and scene generation. 
We first employ a scene encoder to align the language information with the 3D Gaussians, 
resulting in language-augmented 3D Gaussian representations. 
Then, a designed Gaussian projector aligns the 3D Gaussian tokens, 
2D image tokens, and text tokens into a unified latent space. 
Subsequently, a task-aware hybrid sampling strategy is applied 
to select the most relevant 3D Gaussian tokens for the current query, 
which are then fed into the LLM. 
The LLM produces both textual answers and high-level language features 
that encapsulate world knowledge, which are later used 
to guide multi-modal scene generation.
 }
    \vspace{-0.4cm}
    \label{fig:system}
\end{figure*}

\section{Method}
In this paper, we propose GaussianDWM, a unified framework with 3D gaussian scene representation for driving scenarios understanding and generation. The pipeline of our method
is illustrated in Fig.~\ref{fig:system}. The input to our method consists of images $\{I_i\}$, Gaussian ellipsoids $\{G_i\}$, and query text $\{t_i\}$. The framework is composed of three main modules:(i) World tokenizer (Sec.~\ref{sec:world});
(ii) Scene Understanding(Sec.~\ref{sec:understanding});
(iii) Multi-modal Generation (Sec.~\ref{sec:generation}).
We elaborate on the entire pipeline of our system in the following subsections.

\subsection{World Tokenizer}
\label{sec:world}
Our world tokenizer encodes the world observations, i.e., the current multi-view images, into a compact continuous 3D Gaussian representation. We then apply a gaussian projector
to align the selected 3D features to the text space before processing by the VLM.

\noindent \textbf{3D Gaussian Tokenizer}
To preserve texture, 3D structure, and language alignment, we adopt 3D Gaussians as the scene representation for LLM input. We build upon LangSplat~\cite{langsplat} to construct a 3DGS language field, where each Gaussian is augmented with a language embedding $f_i$. These embeddings are obtained from CLIP features, which inherit hierarchical semantics extracted via SAM~\cite{sam}. We then follow the standard 3DGS rendering strategy, incorporating language information directly into the Gaussian primitives. 
\vspace{-4mm}
\begin{equation}
\boldsymbol{F}(v)=\sum_{i \in \mathcal{N}} f_i \alpha_i \prod_{j=1}^{i-1}\left(1-\alpha_j\right)
\end{equation}
\vspace{-5mm}

\noindent where $\boldsymbol{F}(v)$ represents the language embedding rendered at pixel $v$, and $\alpha_i=o_i G_i^{2 D}(v)$. Here $o_i$ is the opacity of the $i$ th Gaussian and $G_i^{2 D}(\cdot)$ represents the function of the $i$-th Gaussian projected onto 2D.
To further reduce memory consumption and improve efficiency, we introduce a scene-wise language autoencoder $E$, which maps the CLIP embeddings $\boldsymbol{F}(v) \in \mathbb{R}^D$ to $\boldsymbol{H}(v)= E\left(\boldsymbol{F}(v)\right) \in \mathbb{R}^d$, where $d \ll D$. We select $d=3,D=512$. Then we learn a decoder $\Psi$ to reconstruction CLIP feature. Our autoencoder can significantly decrease memory requirements while retaining semantic fidelity.

\noindent \textbf{3D Gaussian Projector}
 We first align the extracted 3D Gaussian tokens to the text space. For each Gaussian primitive $G_i$, we represent its attributes as
$G_i = (x_i, o_i, s_i, r_i, f_i)$
where $x_i \in \mathbb{R}^3$ denotes the 3D spatial position, $o_i$ the opacity, $s_i$ the scale, $r_i$ the rotation, and $f_i$ the associated CLIP feature.
For the Gaussian tokenizer, we first apply learnable Fourier embeddings~\cite{nerf} to the 3D coordinates $x_i$:
\vspace{-2mm}
\begin{equation}
    \gamma(x_i) = \left[ \sin(2^k \pi x_i), \cos(2^k \pi x_i) \right]_{k=0}^{L-1},
\end{equation}
\vspace{-6mm}

\noindent where $L$ is set to 10. For the opacity $o_i$, we apply a sigmoid activation $\hat{o}_i = \sigma(o_i)$ to constrain the value to $[0, 1]$. For the CLIP feature $f_i$, we use a pre-trained scene-wise decoder to project it to a 512 dimension $\tilde{f}_i = \Psi(f_i) \in \mathbb{R}^{N\times512}$. $N$ denotes the number of 3D Gaussian ellipsoids. Then, we employ a set of MLP projectors to map Gaussian attribute into a shared 4096-dimensional feature space, i.e., $h_i^x = \phi_x(\gamma(x_i))$, $h_i^{o} = \phi_o(\hat{o}_i)$, $h_i^{s} = \phi_s(s_i)$, $h_i^{r} = \phi_r(r_i)$, and $h_i^{f} = \phi_f(\tilde{f}_i)$, where $\gamma(\cdot)$ is the Fourier embedding and each $\phi_\cdot(\cdot)$ is a learnable MLP. Finally, we fuse the projected features via learnable weights to obtain the Gaussian scene token $\mathcal{G}_i = \sum_{p \in \{x, o, s, r, f\}} \alpha_p \cdot h_i^p$, where each $\alpha_p$ is a trainable scalar normalized by a softmax.
For text queries, we tokenize the input prompts into vocabulary indices and text tokens $\mathcal{T}$ for LLM processing.

\subsection{Scene Understanding}
\label{sec:understanding}
This section introduces the world understanding module. The Large Language Model (LLM) interprets driving scenarios from the world tokenizer outputs $\mathcal{G}_i \in \mathbb{R}^{N\times C}$ according to user instructions. 
Then, the LLM parses the user instruction $\mathcal{T}_i$ and extracts world knowledge from the driving scene, generating both a textual answer $t_i$ and a language feature representation $C_L$, which is later used as a condition signal for scene generation.
 This feature encodes high-level world knowledge as well as spatial information and is later used as a condition for scene generation. We implement the LLM using the widely adopted Qwen3 model~\cite{qwen3}. The overall architecture is:
\vspace{-2mm}
\begin{equation}
    \{t_i,C^l_i\}=LLM(\mathcal{G}_i,\mathcal{T}_i)
\end{equation}
\vspace{-5mm}

\noindent \textbf{Task-aware Language-guided Sampling}
However, directly converting all Gaussians into tokens would exceed the maximum token length limits of LLMs, and the high degree of redundancy in the Gaussian set would make it difficult for the LLM to reason about spatial interactions across different views. To address this, we propose a task-aware hybrid sampling strategy tailored for 3D Gaussian scene representations. For scene understanding tasks, we adopt a global sampling strategy that preserves holistic scene information by selecting a representative subset of Gaussians. We apply both uniform sampling and top-k sampling to select $N = 4096$ gaussian tokens from the hundreds of thousands in the scene, which are then fed into the LLM.
 In contrast, for 2D and 3D visual grounding tasks, we further introduce a language-guided sampling module, which re-tokenizes the dense scene representation into a more compact and sparse form conditioned on the text query. Specifically, we apply similarity calculation between the 3D Gaussian features and the text query to identify and retain only those Gaussians that are most relevant to the query. This design is highly effective for both 2D and 3D grounding tasks, as it enables the model to selectively inject the most relevant 3D spatial information into the language reasoning process. Compared with the previous Hermes framework, our 3D Gaussian--based LLM model can effectively respond to user queries about the driving environment, providing scene descriptions and answers to visual questions. In addition, it supports both 2D and 3D visual grounding tasks, achieving state-of-the-art performance in terms of averaged metrics.

\noindent \textbf{Training Strategy}
Similar to many VLM training protocols, we adopt a two-stage training strategy consisting of an alignment stage and a fine-tuning stage. 
In the first stage, we freeze the entire VLM and train the aligner with full parameters for 5k warm-up steps to align the visual representations with the textual space. 
In the second stage, we further adapt the LLM using LoRA for 30k steps. Both stages share the same training objective.
Following, we use a prefix language modeling, where the model is
conditioned on an input prefix and trained to autoregressively
generate the target continuation:
\vspace{-3mm}
\begin{equation}
\mathcal{L}(\theta,\mathcal{B})=-\sum_{\left\{t_{\text {prefix }}, t_{\mathrm{gt}}\right\} \in B} \sum_{i=1}^{\left|t_{\mathrm{gt}}\right|} \log p_\theta\left(t_{\mathrm{gt}}^{(i)} \mid t_{\mathrm{gt}}^{(<i)}, t_{\text {prefix }}\right),
\end{equation}
where $\theta$ are the model parameters, $\mathcal{B}$ denotes a batch of samples of prefix input $t_{\text {prefix }}$ (text tokens, image tokens and 3D Gaussian tokens), and ground truth response $t_{\mathrm{gt}}. t_{\mathrm{gt}}^{(t)}$ denotes the $t$-th token in the ground truth response sequence.

\subsection{Multi-modal Scene Generation}
\label{sec:generation}
\noindent
In this section, we propose a dual condition multi-modal scene generation model. Our model consists a denoising UNet~\cite{stablediffusion}, and a frozen pre-trained VAE~\cite{vae} to encode RGB images $I_i$, and depth maps $D_i$ into a unified latent space. To satisfy the VAE input specifications, we convert depth maps into pseudo-RGB images via channel replication. The VAE encoding process can be written as:
$
    z_{I}=\mathcal{E}\left(I_i\right), \quad z_{D}=\mathcal{E}\left(D_i\right)
$
During decoding, the VAE decoder $\mathcal{D}$ reconstructs RGB, and depth from the latent variable $z_i$. For the depth map, we average the three output channels of the decoded result to obtain the final single-channel depth prediction:
$
\hat{I}_i=\mathcal{D}\left(z_{I}\right), \quad \hat{D}_i=\frac{1}{3} \sum_{c=1}^3 \mathcal{D}\left(z_{D}\right)_c
$
In the training phase, we random sample on the known camera trajectory, and get the latent code
 of corresponding color and depth images with the VAE encoder $z_I,z_D$. At each timestep $t$, we add noise to the sampled data. Then, we use the projection matrix to project surrounding point cloud at time $t$ to time $t+n$ to serve as low-level image conditions $\{C_I,C_D\}$. Then we concatenate the noisy latent representations of each modality with the low-level image control signal and high-level language control signal $C_L$ from LLM as the input to the denoising diffusion network. 
The model is using a v-prediction
objective. The target $\mathbf{v}_t$ is defined as:$\mathbf{v}_t=\alpha_t \boldsymbol{\epsilon}_t-\sigma_t d_t,$
where $\boldsymbol{\epsilon}_t \sim \mathcal{N}(0, I)$ denotes the sampled gaussian noise, 
$\alpha_t$ and $\sigma_t$ represent the time-dependent noise scheduling coefficients, and $\boldsymbol{d}_t$ corresponds to the noisy input modality requiring denoising. The training objective is defined as:
\vspace{-2mm}
\begin{equation}
\mathcal{L}=\mathbb{E}_{d, \boldsymbol{\epsilon}, t, \boldsymbol{s}}\left\|\mathcal{F}_\theta\left(d_t, d_{\mathrm{ref}}, C_{I}, C_{D}, C_{L},\boldsymbol{s}\right)-\mathbf{v}_t\right\|_2^2,
\end{equation}
\vspace{-6mm}

\noindent The low-level conditions $C_{I}$ and $C_{D}$ represent 
the scene’s texture and geometric information, guiding the generation process. 
Meanwhile, the high-level language feature $C_{L}$ encapsulates comprehensive 
world knowledge extracted from the LLM. 
By conditioning the generation on multiple levels of information, 
our model achieves more accurate and consistent temporal and spatial synthesis.

For spatial generation, we project the surrounding point cloud 
using the spatial transformation of the query frame to obtain a sparse condition map. 
For temporal generation, we utilize the trajectory predicted by the front-end LLM 
to project the point cloud and construct a temporal sparse condition map, 
enabling temporally coherent scene generation.

Our generation model supports both spatial scene generation, i.e., novel view synthesis with spatial shifts of $1\text{m}$ or $2\text{m}$, and temporal scene generation, i.e., future scene prediction at $1\text{s}$ and $2\text{s}$ into the future.

\section{Experiments}
\begin{table*}[htbp]
\centering
\resizebox{\textwidth}{!}{%
\begin{tabular}{llcccccccccccc}
\hline
 &  &  & \multicolumn{3}{c}{2D RD \& Pre $\uparrow$} & \multicolumn{3}{c}{2D VG $\uparrow$} & \multicolumn{3}{c}{3D VG $\uparrow$} & Plan $\uparrow$ &  \\ \cline{4-13}
\multirow{-2}{*}{Model} & \multirow{-2}{*}{Years} & \multirow{-2}{*}{LLM} & BLEU & Rouge\_L & CIDEr & mAP & F1 & MIoU & Pr & mAP & F1 & Acc & \multirow{-2}{*}{Avg. $\uparrow$} \\ \hline
LLaVA1.5 & 2024 & Vicuna-7B & 64.23 & 76.69 & 74.82 & 0.10 & 0.16 & 14.31 & 6.51 & 5.33 & 3.12 & 36.20 & 28.16 \\
MiniCPM-V 2 & 2024 & MiniCPM-2B & 47.43 & 63.16 & 69.88 & 0.11 & 0.13 & 13.34 & 0.97 & 1.55 & 0.86 & 36.69 & 23.41 \\
MiniCPM-V 2.6 & 2024 & QWen2-7B & 47.92 & 69.11 & 70.20 & 0.36 & 0.49 & 18.74 & 1.97 & 1.61 & 0.93 & 36.42 & 24.78 \\
InternVL1.5-2B & 2024 & InternLM2-7B & 67.14 & 81.10 & {\ul 79.83} & 14.74 & 17.64 & 55.43 & 28.05 & 21.73 & 12.92 & 53.96 & 43.25 \\
InternVL1.5-4B & 2024 & Phi3-4B & 66.63 & 80.64 & 79.24 & 14.27 & 17.60 & 53.52 & 25.14 & 19.46 & 11.63 & 40.25 & 40.84 \\
QWen2VL & 2024 & Qwen2-2B & 67.92 & 80.24 & 78.51 & 17.11 & 20.87 & 57.24 & 12.82 & 10.20 & 6.12 & 45.59 & 39.66 \\
Qwen2VL & 2024 & QWen2-7B & 66.65 & 78.57 & 77.97 & 16.06 & 20.04 & 55.51 & 20.64 & 16.26 & 9.82 & 49.33 & 41.09 \\
InternVL2-1B & 2024 & Qwen2-0.5B & 66.89 & 81.00 & 79.59 & 16.70 & 20.21 & 55.94 & 23.36 & 18.35 & 10.94 & 44.08 & 41.71 \\
InternVL2-2B & 2024 & InternLM2-2B & 66.77 & 80.87 & 79.62 & 16.12 & 19.49 & 55.29 & 27.83 & 21.09 & 12.58 & 44.61 & 42.43 \\
InternVL2-4B & 2024 & Phi3-4B & 66.88 & 80.76 & 79.29 & 19.14 & 23.47 & 59.07 & 25.28 & 20.12 & 11.97 & 40.43 & 42.64 \\
InternVL2-8B & 2024 & InternLM2.5-7B & 67.32 & \textbf{81.39} & \textbf{80.01} & {\ul 20.61} & {\ul 25.24} & {\ul 61.90} & 31.47 & 24.67 & 14.70 & 46.93 & 45.42 \\
DriveMonkey & 2025 & InternLM2.5-7B & {67.50} & {\ul 81.15} & 79.79 & 19.47 & 24.02 & 59.36 & 51.90 & {\ul 34.53} & 20.86 & \textbf{82.64} & {\ul 52.12} \\ \hline
Bevformer & 2022 & - & \multicolumn{6}{c}{} & 44.50 & 23.69 & 1.576 & \multicolumn{2}{c}{} \\
PETR & 2022 & - & \multicolumn{6}{c}{} & \textbf{55.80} & 31.34 & 20.58 & \multicolumn{2}{c}{} \\
CAPE & 2023 & - & \multicolumn{6}{c}{\multirow{-3}{*}{Unsupported}} & {\ul 55.02} & 32.94 & {\ul 21.33} & \multicolumn{2}{c}{\multirow{-3}{*}{Unsupported}} \\ \hline
\rowcolor[HTML]{CBCEFB} 
\noalign{\vskip 2pt}
GaussianDWM & 2025 & Qwen3-8B & \textbf{68.78} & 81.06 & 78.72 & \textbf{34.95} & \textbf{40.49} & \textbf{71.85} & 50.66 & \textbf{52.78} & \textbf{32.05} & {\ul 80.95} & \textbf{59.23} \\ 
\noalign{\vskip 2pt}
\hline
\end{tabular}%
}
\caption{The comparison between our \textbf{GaussianDWM} and other state-of-the-art models on the NuInteract dataset~\cite{drivemonkey}. 
The scene understanding task includes four subtasks: region description and perception, 2D visual grounding, 3D visual grounding, and planning. 
Our method achieves state-of-the-art average performance across all four tasks, which fully demonstrates the effectiveness of introducing a 3D Gaussian scene representation for enhancing the LLM’s capability to understand 3D spatial information.
 }
\label{tab:vqa}
\end{table*}

\begin{table*}[htbp]
\centering
\resizebox{\textwidth}{!}{%
\begin{tabular}{cccccccccccccc}
\hline
\multirow{2}{*}{Finetune} & \multirow{2}{*}{Gaussian} & \multirow{2}{*}{Sampling} & \multicolumn{3}{c}{2D RD \& Pre $\uparrow$} & \multicolumn{3}{c}{2D VG$\uparrow$} & \multicolumn{3}{c}{3D VG$\uparrow$} & Plan$\uparrow$ & \multirow{2}{*}{Avg.$\uparrow$} \\ \cline{4-13}
 &  &  & BLEU & Rouge\_L & CIDEr & mAP & F1 & MIoU & Pr & mAP & F1 & Acc &  \\ \hline
zeroshot & w/o &  & 2.91 & 12.68 & 0.59 & 0.00 & 0.00 & 12.24 & 48.75 & 47.59 & 29.12 & 0.00 & 15.39 \\
finetuned & w/o &  & 65.09 & 78.35 & 76.56 & 30.01 & 35.45 & 69.01 & 50.36 & 51.88 & 31.43 & 45.09 & 53.32 \\
finetuned & w & Random &  { \ul 66.19} & {\ul 79.00} & {\ul 76.97} & {\ul 33.94} & {\ul 39.37} & {\ul 71.40} & {\ul 50.94} & {\ul 52.85} & {\ul 32.03} & {\ul 49.43} & 55.21 \\
finetuned & w & Top-k + Uniform & \textbf{68.78} & \textbf{81.06} & \textbf{78.82} & 33.89 & 39.31 & 71.37 & \textbf{51.16} & \textbf{52.87} & \textbf{32.05} & \textbf{80.95} & {\ul 58.93} \\
\rowcolor[HTML]{CBCEFB} 
\noalign{\vskip 2pt}
finetuned & w & Top-k + Uniform + similarity & \textbf{68.78} & \textbf{81.06} & \textbf{78.82} & \textbf{34.95} & \textbf{40.49} & \textbf{71.85} & 50.66 & 52.78 & \textbf{32.05} & \textbf{80.95} & \textbf{59.23} \\ \hline
\noalign{\vskip 2pt}
\end{tabular}%
}
\vspace{-0.2cm}
\caption{ We conduct ablation studies to validate the effectiveness of each proposed component, including the 3D Gaussian scene representation, the top-k and uniform sampling strategies, and the similarity–based sampling module. Note that the similarity sampling strategy is applied only to grounding tasks requiring focused attention (e.g., 2DVG and 3DVG).} 
\label{tab:ablation1}
\vspace{-0.4cm}
\end{table*}

\subsection{Dataset and Evaluation Metric}
\noindent
We evaluate our method on two datasets for both scene understanding and scene generation. NuScenes~\cite{nuscenes} is a widely used autonomous driving dataset. We use six surrounding camera images as inputs. NuInteract~\cite{drivemonkey}  are recent benchmark for scene understanding. NuInteract provides $\sim$1.5M annotations and supports multiple tasks, including 2D perception and 3D visual grounding.  For scene understanding, we adopt standard language and captioning metrics, including gCIDEr~\cite{cider}, BLEU-4~\cite{bleu}, ROUGE~\cite{rouge}. For scene generation, we use FID for images, and FVD for videos as evaluation metrics.
\subsection{Implementation Details}
Our training pipeline consists of three stages. In the first stage, we train the Gaussian tokenizer, projector, and the proposed sampling strategy independently. We then integrate these components with the LLM and perform joint fine-tuning. All models in this stage are trained using 16 NVIDIA A100 GPUs. In the second stage, we train the multi-modal generation module. We start by training a low-resolution ($224 \times 400$) RGB video generation model, extend it to low-resolution RGB-D generation, and finally refine it into a high-resolution ($424 \times 800$) RGB-D video generator using a mixed-frame-length strategy. The model is optimized using simulation-free rectified flow and a v-prediction loss~\cite{vprediction}. In the final stage, we perform end-to-end joint optimization over all components to ensure consistency between scene understanding and scene generation.

\begin{figure*}[h]
    \centering
    \includegraphics[width=\linewidth]{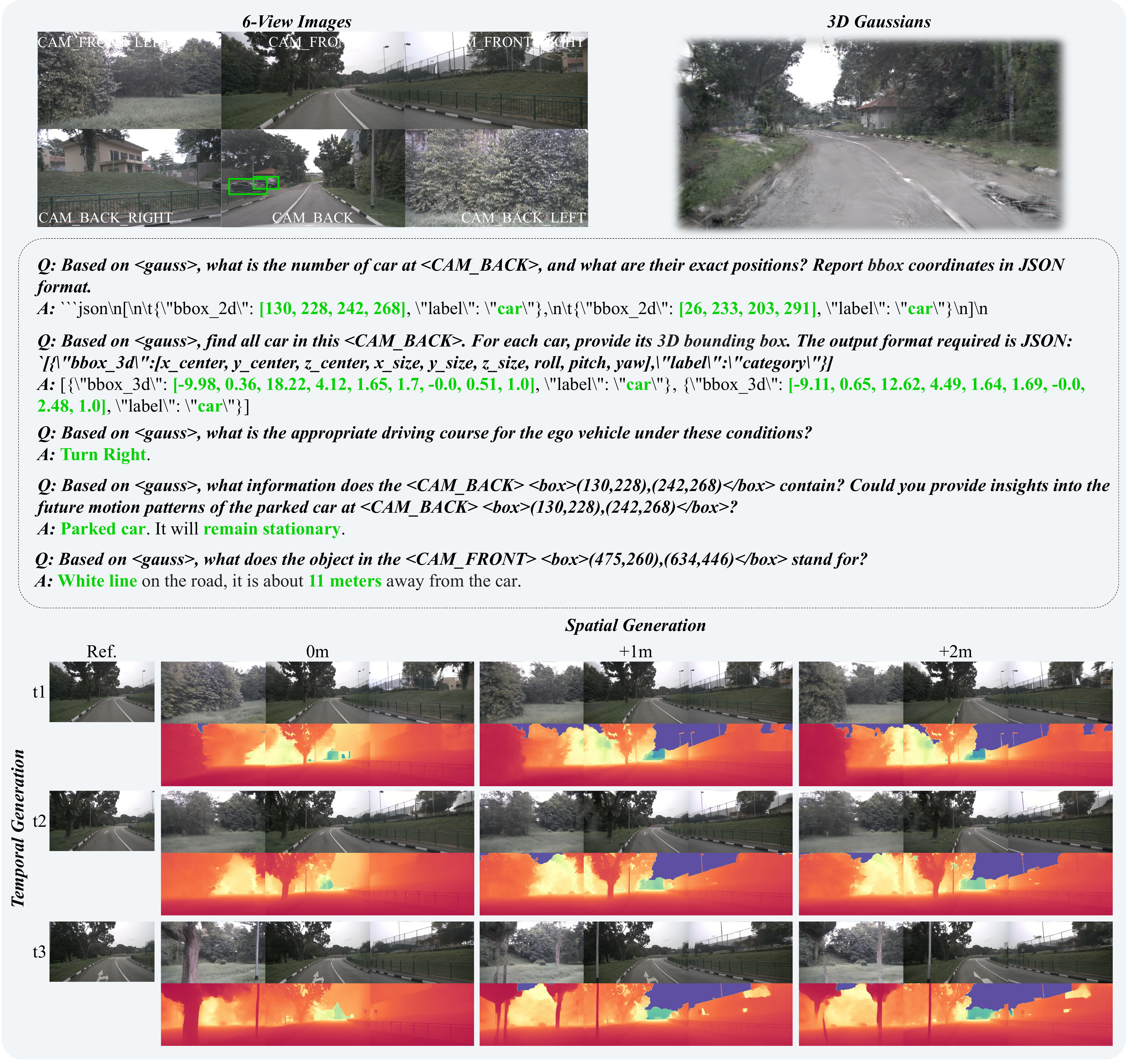}
    \vspace{-0.2cm}
    \caption{\textbf{Qualitative results for scene understanding and scene generation}. From top to bottom, we display the multi-view input of the
current scene and the 3D Gaussian ellipsoids, the scene understanding results, and the spatial and temporal scene generation results.
 }
    \vspace{-0.4cm}
    \label{fig:experiments}
\end{figure*}

\subsection{Scene Understanding}
In this section, we present the results of scene understanding, which are divided into four tasks: region description and perception, planning, 2D visual grounding, and 3D visual grounding. 
The results are shown in Tab.~\ref{tab:vqa}. 
The experimental settings strictly follow Drivemonkey~\cite{drivemonkey}. 
We primarily compare two categories of methods: (1) LLM-based visual grounding approaches such as LLaVA~\cite{LLaVA} and InternVL~\cite{internvl}, and (2) specialized 3D detection models such as BEVFormer~\cite{bevformer}, PETR~\cite{petr}, and CAPE~\cite{cape}. 
As shown in Tab.~\ref{tab:vqa}, GaussianDWM significantly outperforms previous LVLMs in terms of average metrics across all tasks on the NuInteract test dataset. 
It achieves state-of-the-art performance in terms of averaged metrics on the RDP, 2D VG, 3D VG, and Planning tasks, outperforming previous state-of-the-art methods by a relative margin of 13.6\%, effectively demonstrating the importance of introducing a 3D Gaussian scene representation for enabling LLMs to better comprehend 3D spatial information. 
Compared with previous point cloud– and BEV-based scene representations, our 3D Gaussian formulation allows for explicit alignment between 3D structural, texture, and language features, leading to more efficient and semantically consistent information injection into LLMs. 
Moreover, our task-aware hybrid sampling strategy efficiently selects Gaussian tokens most relevant to the query text while filtering out redundant 3D Gaussians. 
Compared with the current state-of-the-art VQA method Drivemonkey, our model shows clear advantages in both 2D and 3D visual grounding tasks, while also achieving comparable performance to specialized 3D detectors designed specifically for 3D VG.

\subsection{Scene Generation}
In this section, we present the evaluation of multi-modal scene generation, which includes both spatial scene generation and temporal scene generation. 
All experiments are conducted on the nuScenes dataset~\cite{nuscenes}. 
Following previous works~\cite{dist4d,magicdrive}, we interpolate the 2Hz keyframe annotations to a higher frame rate of 12Hz, and the experimental settings strictly follow~\cite{dist4d}. 

\begin{figure*}[h]
    \centering
    \includegraphics[width=\linewidth]{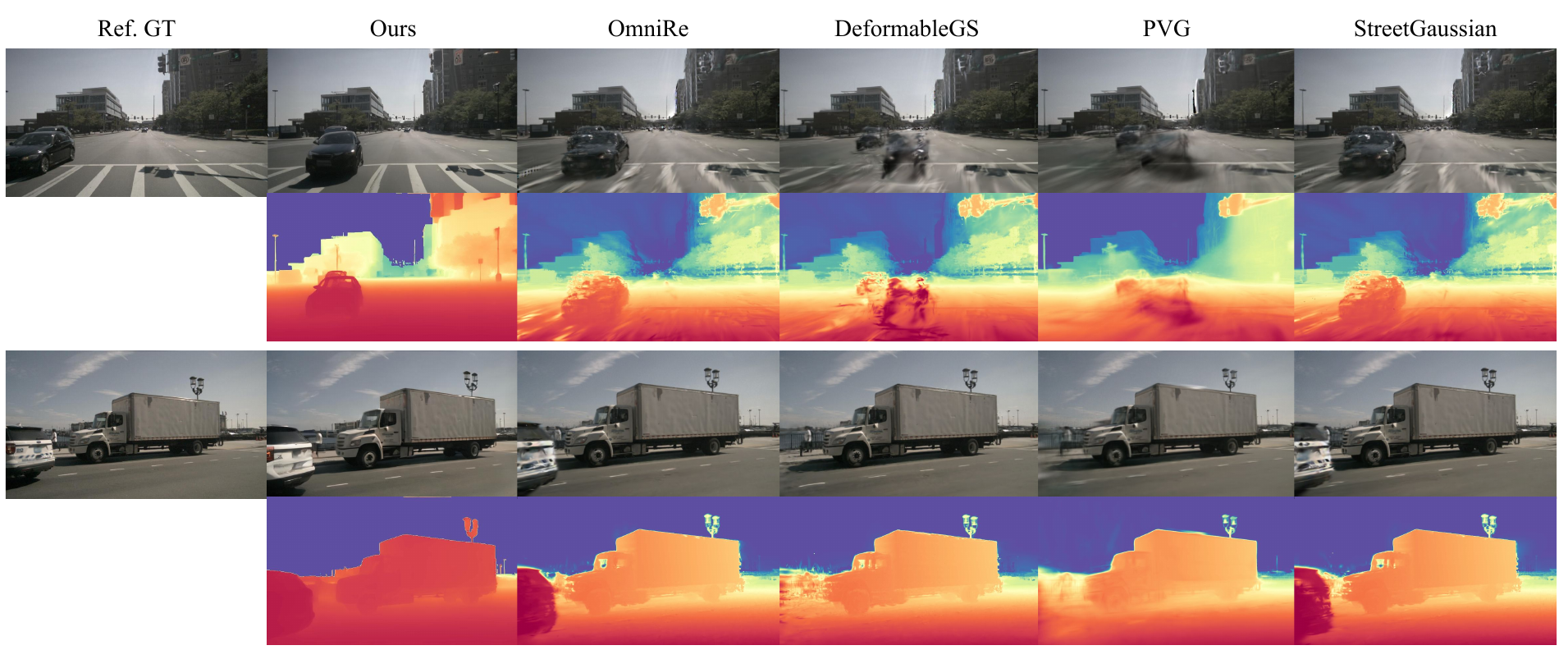}
    \vspace{-0.2cm}
    \caption{ \textbf{Qualitative comparison of RGB-D NVS with 2m shift}. Compared with state-of-the-art reconstruction-based methods for spatial NVS \cite{omnire, deformable3dgs, pvg, streetgaussian}, our method reduce artifacts of dynamic objects and preserves temporal-spatial consistency across large viewpoint shifts. 
 }
    \vspace{-0.4cm}
    \label{fig:generation1}
\end{figure*}

We compare our method with representative street-view synthesis approaches, including PVG~\cite{pvg}, EmerNeRF~\cite{emernerf}, StreetGaussian~\cite{streetgaussian}, OmniRe~\cite{omnire}, FreeVS~\cite{wang2025freevs}, DiST-4D~\cite{Guo2025ICCV}. 
As shown in Tab.~\ref{tab:spatialgeneration}, our method outperforms all existing methods and achieves state-of-the-art performance. 
We use FID and FVD as evaluation metrics since ground-truth data is not available after spatial shifts. 
Our method demonstrates superior consistency and photorealism under extreme viewpoint shifts ($\pm 4$m), outperforming direct reconstruction-based approaches. 
This indicates that our framework effectively combines the advantages of 3D Gaussian scene representation and diffusion-based generative modeling. 
Under the guidance of our dual-condition mechanism—leveraging both high-level world knowledge and low-level geometric cues—our method achieves state-of-the-art spatial and texture fidelity under large viewpoint variations. 

We also visualize the qualitative results of spatial scene generation in Fig.~\ref{fig:experiments}. We further compare our method with several existing novel view synthesis approaches, and the qualitative results are presented in Fig.~\ref{fig:generation1}. 
As shown, our RGB and depth generations exhibit photorealistic quality, demonstrating the complementary strengths of world knowledge, diffusion-based generation, and 3D Gaussian scene representation.

\begin{table}[]
\centering
\setlength{\tabcolsep}{0.9mm}
\scalebox{0.85}{
\begin{tabular}{lcccccc}
\toprule
\multirow{2}{*}{Method} & \multicolumn{2}{c}{Shift $\pm$ 1} & \multicolumn{2}{c}{Shift $\pm$ 2} & \multicolumn{2}{c}{Shift $\pm$ 4} \\ \cmidrule(lr){2-3} \cmidrule(lr){4-5} \cmidrule(lr){6-7}  
                        & FID $\downarrow$            & FVD $\downarrow$             & FID $\downarrow$             & FVD $\downarrow$            & FID $\downarrow$            & FVD $\downarrow$            \\ \midrule
PVG                     & 48.15           & 246.74           & 60.44            & 356.23          & 84.50           & 501.16          \\
EmerNeRF                & 37.57           & 171.47           & 52.03            & 294.55          & 76.11           & 497.85          \\
StreetGaussian          & 32.12           & 153.45           & 43.24            & 256.91          & 67.44           & 429.98          \\
OmniRe                  & 31.48           & 152.01           & 43.31            & 254.52          & 67.36           & 428.20          \\
FreeVS                  & 51.26           & 431.99           & 62.04            & 497.37          & 77.14           & 556.14          \\
DiST-S                  & 10.12           & 45.14            & 12.97            & 68.80           & \textbf{17.57}  & \textbf{105.29} \\ \midrule
Ours                    & \textbf{8.36}   & \textbf{44.50}   & \textbf{11.27}   & \textbf{68.17}  & 18.81           & 116.40          \\ \bottomrule
\end{tabular}}
\caption{ \textbf{Quantitive comparison} between our method and other novel view synthesis methods on nuScenes dataset.}
\label{tab:spatialgeneration}
\vspace{-0.5cm}
\end{table}


The proposed dual-condition design enables high-level world knowledge to effectively guide the model’s temporal reasoning for future scene prediction, while the low-level sparse condition constrains 2D texture and style consistency. 
This demonstrates the importance of incorporating explicit \textbf{world knowledge} into world models to improve both semantic and spatial coherence during temporal generation. 
Overall, GaussianDWM is capable of simultaneously understanding and generating complex driving scenarios, highlighting a promising research direction toward unified scene understanding and generation in world modeling.

\subsection{Ablation Study}
In this section, we conduct comprehensive ablation studies to validate the effectiveness of each component in our proposed framework. 
We systematically analyze the impact of the key modules, including the 3D Gaussian scene representation, the hybrid sampling strategies, and the dual-condition generation design, to demonstrate their respective contributions to the overall performance.

\noindent \textbf{3D Gaussian Representation}
Compared with other scene representations, the 3D Gaussian scene representation provides a better ability to encode environmental features, including both texture and geometric information, and can be directly aligned to the 3D space through language features. The results are shown in Tab.~\ref{tab:ablation1}. Through ablation studies, we fully verify that the introduction of 3D Gaussian representations enhances the LLM's 3D scene understanding capability, enabling it to better perceive the geometric and semantic information of the scene.

\noindent \textbf{Hybrid Sampling}
Based on the observed redundancy in the 3D Gaussian scene representation, we design a hybrid sampling strategy to extract the most informative components of the environment. The results are shown in Tab.~\ref{tab:ablation1}. By comparing random sampling, uniform+top-k sampling, and similarity-based sampling, we observe that when similarity-based sampling is applied between the text query and the 3D Gaussian tokens, the model can more effectively select task-relevant Gaussian tokens, leading to improved scene understanding performance.

\noindent \textbf{High-level World Knowledge}
Compared with other world model and scene generation frameworks, we design a dual-condition guidance framework that controls scene generation at different levels. The ablation results for this part are shown in Tab.~\ref{tab:world}. It can be observed that our method performs better in long-sequence prediction and under large viewpoint variance, demonstrating the effectiveness of incorporating world knowledge.
\vspace{-0.2cm}

\begin{table}[]
\setlength{\tabcolsep}{0.9mm}
\centering
\scalebox{0.85}{
\begin{tabular}{ll|ll}
\hline
Low-Level & High-level & FID$\pm$1m & FID$\pm$2m \\ \hline
\ding{55}      & $\checkmark$       & -       & -       \\
$\checkmark$     & \ding{55}        & 10.12      & 45.14      \\ \hline
$\checkmark$     & $\checkmark$       & \textbf{8.36}       & \textbf{44.5}       \\ \hline
\end{tabular}}
\caption{\textbf{Ablation Study of dual-condition generation mechanism}. “–” denotes failure under the setting.}
\label{tab:world}
\vspace{-0.5cm}
\end{table}

\section{Conclusion}
\vspace{-0.1cm}
This paper introduces a novel unified driving world model framework that supports both scene understanding and generation within a single architecture. We bridge the gap between these two tasks by proposing a new 3D Gaussian scene representation combined with a task-aware hybrid Gaussian sampling strategy, enabling effective world-query injection into the LLM.
Extensive experiments validate the effectiveness of the proposed framework, demonstrating significant improvements in both scene understanding and generation. We believe this work an important step toward unified driving world models with 3D Gaussian scene representations.

\textbf{Acknowledgement}
This work was supported by National Key R\&D Program of China (Grant No.2024YFB4708900). It was also supported in part by the Natural Science Foundation of
China under Grant 62225309,U24A20278, 62361166632. We sincerely thank Yikang Ding from Kuaishou Kling AI for his kind help and support.


{
    \small
    \bibliographystyle{ieeenat_fullname}
    \bibliography{main}
}


\end{document}